# ExKG-LLM: Leveraging Large Language Models for Automated Expansion of Cognitive Neuroscience Knowledge Graphs

**Ali Sarabadani**
Department of Computer Engineering and Information Technology, University of Qom, Qom, Iran.
a.sarabadani@stu.qom.ac.ir

**Kheirolah Rahsepar Fard**
Department of Computer Engineering and Information Technology, University of Qom, Qom, Iran.
rahsepar@qom.ac.ir

**Hamid Dalvand**
Department of Occupational Therapy, School of Rehabilitation, Tehran University of Medical Sciences, Tehran, Iran.
hamiddalvand@gmail.com

## Abstract

**Objective:** This paper introduces ExKG-LLM, an innovative framework designed to automate expanding cognitive neuroscience knowledge graphs (CNKG) using large language model (LLM). This model includes increasing KGs' accuracy, completeness and usefulness in cognitive neuroscience.

**Method:** To address the limitations of existing tools for creating knowledge accounts, this is especially true in dealing with the complex hierarchical relationships within the cognitive neuroscience literature. We use a large dataset of scientific paper and clinical reports, the ExKG-LLM framework, new entities and relationships in CNKG to apply state - state of the art LLM to extract, optimize and integrate, evaluating performance based on metrics such as precision, recall and graph density.

**Findings:** The ExKG-LLM framework achieved significant improvements, including precision of 0.80 (an increase of 6.67%), recall of 0.81 (an increase of 15.71%), F1 score of 0.805 (an increase of 11.81%), and number of edge nodes increased by 21.13% and 31.92%, respectively. Also, the density of the graph decreased slightly. Reflecting the broader but more fragmented structure, engagement rates have also increased by 20%, highlighting areas where stability needs improvement. From the perspective of a complex network, increasing the diameter of CNKG to 15 compared to 13 shows that although the size of ExKG-LLM has increased, more steps are now required to discover additional nodes. Although time complexity improved to O($n \log n$), space complexity became less efficient, rising to O($n^2$), indicating higher memory usage for managing the expanded graph.

**Conclusion:** ExKG-LLM shows significant potential to improve knowledge generation and expansion. Especially in cognitive neuroscience, CNKGs are expanding with increasing edge nodes. Improved grouping Provides powerful resources for applications such as semantic search, Data-driven research and clinical decision-making in neurological disorders from a complex network perspective. While increasing diameter indicates a more distributed structure is possible. While the observed strong local clustering ensures that the graph remains navigable and useful for complex queries, the framework is also adaptable to a wider scientific field. It positions ExKG-LLM as a multi-purpose tool for KG technology in various fields.

# 1.Introduction

Cognitive neuroscience is an interdisciplinary field that combines insights from psychology. Neuroscience and computer science are essential in revealing the neural mechanisms underlying human cognition. A structured representation of the entities within a specific domain and their relationships is presented. These graphs allow for systematic data storage. Retrieving and exploring complex information gives researchers a powerful way to navigate the complex networks of cognitive neuroscience.

Despite its many functional benefits, creating high-quality KG has traditionally been labour-intensive and time-consuming, especially in a field as complex as cognitive neuroscience. This requires careful management of the complexity of the relationships and the sheer volume of data available. However, recent advances in natural language processing (NLP) and the development of LLM have opened up the possibility. New for automatic KG generation and extension. Using LLM's deep language understanding capabilities, automating knowledge extraction and integration from ample text resources is now possible and enhancing the accuracy and breadth of KGs.

The important thing about developing a KG is "Link Prediction", which attempts to infer missing connections between entities based on the existing graph structure. Link prediction in cognitive neuroscience can uncover new connections between cognitive function, neural domains, and disorders. It can add additional value and lead to a deeper understanding of brain function and cognition.

In addition to link prediction, considering CNKG as a complex network allows for in-depth analysis of its structural properties indicators such as average clustering coefficient and diameter. It provides insight into how well nodes are grouped and how efficiently the information flows in the network, and such analysis is important for understanding the connectivity and behavior of CNKG, especially as the graph expands.

Developing a comprehensive KG like CNKG is critical to improving our understanding of brain function and cognition. Automatically extending CNKG using LLM reduces the time and effort required to maintain and upgrade the date graph And guarantees that the graphs are up to date with the latest scientific research. This automated method improves the scope and accuracy of CNKG, making it an invaluable tool for research and clinical applications.

The ExKG-LLM framework implemented in this study represents significant progress in automatic KG expansion. Using state-of-the-art LLM, ExKG-LLM integrates micro-level relationships from extensive datasets in CNKG. This approach increases the depth and breadth of the knowledge map. This ensures that it reflects the most up-to-date and relevant information in the field. The improvement of important parameters, precision, recall, graph density, and complex network features, such as clustering coefficients and diameter, increase in value, provide the potential to reveal previously unknown connections and increase the understanding of the nervous system as a result of the implementation of this model.

The expansion of KGs in complex domains such as cognitive neuroscience offers numerous advantages. These graphs systematically extract and structure complex information from scientific texts, helping researchers access existing knowledge more efficiently and uncover hidden relationships between concepts. Additionally, leveraging LLM for the automatic expansion of KGs ensures that new data is quickly integrated, keeping the graph up-to-date. This not only enhances the accuracy and breadth of the graph but also increases its applications in clinical decision-making and scientific research. Given the rapid growth of information in scientific fields, automatic expansion of these graphs is essential to maintain the coherence and completeness of knowledge, facilitating the discovery of new patterns and connections between entities.

This paper is structured as follows: Section 2 reviews related work, Section 3 details the methodology, Section 4 presents evaluation results, and Section 5 discusses the findings and their implications for future research in cognitive neuroscience.

## 2. Related work

PolarisX, an auto-evolving KG that will automatically augment itself with information gleaned from news sites and social media, is introduced in this paper [1]. It mainly focuses on neologisms. By using the multilingual BERT model, this system is built to extract new relationships and incorporate them into the knowledge graph to follow all current changes within languages and introduce new concepts without language dependency. There are two main innovations in this knowledge graph: collecting neologisms and being independent of language in data collection.

In this paper [2], the authors propose exploring drug indication expansion through the use of knowledge graphs. The principal aim is to find new therapeutic uses for pre-existing drug targets. They discuss some methods along with hybrid knowledge graphs that combine gene data with scientific literature and show features of relationships among genes, disease, and tissues. Then, they test new prediction methods on shortest paths and embeddings, showing a significant improvement in prediction accuracy by integrating data regarding tissue expression. As it is transparent and interpretable, this knowledge graph can be used very effectively in drug indication expansion.

According to [3], the idea of KG applications is directed toward biomedical research. Knowledge graphs can automatically extract biomedical textual information and frame rules and apply label functions to identify associated relationships among biological entities such as genes and diseases. This research hastens the process of creating label functions, and at the same time, an advanced enriched up-to-date knowledge graph is constructed that could benefit pharmaceutical and biomedical research.

The KGs in healthcare have been created from various sources largely during the time of the pandemic, as in the case with COVID-19, in order to develop treatment procedures for the virus [4]. Such KGs have been equally found to be vital instruments in clinical decision support [5] and in discovering false health information, thus making them yet of great value to the healthcare sector [6]. By collating data from various domains, it has thus been possible to create such knowledge graphs that are more comprehensive and thus enable drawing of many meaningful conclusions in medical research and health care.

J. Xu et al. constructed knowledge graph using multitudes of integrated multi-source commons from infrastructures: PubMed literature, ORCID, MapAffil, and NIH ExPORTER project data [7]. The PubMed knowledge graph was used to link bio-entities drawn from 29 million PubMed abstracts through a BioBERT-based Named Entity Recognition (NER) model with entities extracted from multiple sources. Unique authors from ORCID and MapAffil as well as NIH ExPORTER funding data were incorporated for building the robust network of biomedical information [7]. It thus allows a better understanding of the scientific entity relationship and further discovery in biomedicine.

At present, researchers have found a way to integrate LLMs in the construction and expansion of the healthcare and medical KGs . For instance, the DIAMOND-KG developed by Alharbi et al. is a medical context-aware drug indication KG. The study showed that around 71% of indications were related to at least one medical context. So it raises importance of these contextual information in improving the quality of KGs[8]. Furthermore, Mann et al. discussed a new framework for treatment discovery based on known symptoms or diseases. The new model consists of a generalized LLM which can learn from all existing knowledge sources available in the medical field and form a complete KG [9].

In one study, Wu et al. postulated reguloGPT, a GPT-4-based entity that is exceptionally competent at harvesting named entities, harvesting N-ary relationships, and predicting the context of molecular regulatory sentences. The model is developed to build KGs around molecular regulatory pathways[10]. On the contrary, Jiang et al. leveraged the power of patient-specific KGs and used them with a Bidirectional Attention-Enhanced Graph Neural Network (BAT GNN) that helped bring a change to medical decision-making and even diagnosis in the patient context settings by using LLMs [11].

Likewise, Xu et al. offered RAM-EHR, which translates knowledge sources into much-text format via retrieval-augmented multimodal means to be used in better medical prediction from electronic health records (EHR). Such integration between LLMs and KGs improves clinical decision-making [12]. Gao et al. were successful in the development of a diagnostic reasoning system, named DR.KNOWS, which employs KGs with the Unified Medical Language System (UMLS) and a clinical diagnosis model to improve diagnostic accuracy and interpretability [13].

Finally, Zhu et al. developed the REALM model, a novel approach to integrating clinical notes and multivariate time-series data using Retrieval-Augmented Generation (RAG) technology. This combination further improves diagnostic performance in complex medical cases by synergistically employing LLMs with KGs [14]. These studies together present a picture of increasing roles for LLMs in constructing and expanding medical KGs for advances in healthcare research and clinical decision-making. Numerous studies are now focusing on improving and developing medical KGs (peacekeeping) LLM. Wu T. et al. have reviewed the progress on Chinese medical KGs and the contributions of LLMs on the construction and improvement of these graphs within the field in detail [15]. Agrawal et al. investigate the feasibility of using LLMs for zero-shot and few-shot information extraction from clinical text and demonstrate how it could be used on the models such as InstructGPT to extract relevant medical information with the lowest training burden [16,17].

Frisoni et al. have compared the performance of a graph-based transformer and pre-trained LLMs, such as T5 and BART, with respect to the graph-to-text and text-to-graph tasks. Their observations yielded mixed results, showing that some models perform better in certain contexts vis-à-vis biomedical applications . On the other hand, Choudhary and Reddy made some investigations into complex logical reasoning over the KG by using LLMs and found that the capabilities of such models provide for an intricate reasoning task based on graph data[19] .

In recent years, the construction of medical KGs has received attention to enhance the decision-making process in healthcare. Previous studies have concentrated on building knowledge graphs, which intuitively show the relationships between medical concepts to make the user experience be more user-friendly in retrieval of medical information [20]. As an example, Shanghai Shuguang Hospital built a KG for traditional Chinese medicine without success in automatically prescribing drugs for the patient [21]. Automating clinical prescription, for example, has been accomplished in one of the recent advancements in this area through the use of LLMs for building medical KGs, as these models, through prompt engineering, have automated most steps in the construction of KGs, and have even been used to develop knowledge bases for the identification of certain conditions, for instance, autism spectrum disorder [22].

They have additionally begun using LLMs for more and more medical information extraction purposes. These models are successfully identifying medical entities and extracting relationships between entities and disambiguating them as well [23]. They prove to be an effective tool in KG completion rather than merely identification and relationship extraction. For instance, link predictions for KGs using ChatGPT have shown significant improvement in the prediction accuracy of medical information [24]. Finally, these methods will improve the quality of diagnosis and treatment in clinics.

This paper [25] proposes new ways of developing a heart failure KG using LLMs. It employs a method that combines prompt engineering with the TwoStepChat for improved performances compared to other BERT-based models, especially with regards to out-of-distribution OOD information. Moreover, this method greatly shortens the amount of time spent on obtaining medical information extraction.

### 3.Methodology

Methodologically, we first collect and process cognitive neuroscience data. Section 3.1 is data collection and preprocessing, which includes data preparation through tokenization, normalization, and entity identification. In the next step, section 3.2, we will use the LLM to extract and combine new entities and relations compatible with the existing knowledge maps to improve the structure by presenting and expanding the knowledge maps.

### 3.1.Data Collection and Preprocessing

Among the data used in this research are the collection of scientific articles, clinical reports and other sources related to cognitive neuroscience. Emphasis on stroke and neurological disorders are the keywords of data collection. This dataset was drawn from various open repositories, including the PubMed clinical trials database and domain-specific medical journals. These resources have been selected to ensure coverage of a broad range of topics in comprehensive cognitive neuroscience. These sources are considered sources of different information for extracting knowledge. The dataset is continuously updated to incorporate the latest research findings and clinical developments. After that, pre-processing steps are carefully applied to prepare the raw text data for analysis. Preprocessing steps are applied to the data in the following order[26]:

1. **Tokenization:** The initial stage divides textual data into sentences and their component words. This tokenization process is performed using advanced natural language processing (NLP) toolkits such as SpaCy and NLTK, which accurately segment text while maintaining the semantic integrity of sentences.

2. **Normalization:** After tokenization comes text normalization, which involves converting all text to lowercase to ensure consistency. It removes punctuation marks that do not affect the semantic meaning and uses word division to reduce words to their root form.

3. **Named Entity Recognition (NER):** This phase uses a detailed pre-trained named entity recognition (NER) model for biomedical-clinical text to identify and extract important entities related to neurocognitive disease names among these entities: disease names (e.g., "stroke", "Alzheimer's disease"), biomarkers (e.g., "APOE4", "amyloid-beta"), and treatment methods (e.g., "thrombolysis", "cognitive therapy"). NER models leverage domain-specific integration to ensure high accuracy and recall entity separation reduces false positives and saves sensitive biomedical terminology.

4. **Dependency Parsing:** Dependency analysis is performed to clarify syntactic and semantic relationships between identified entities. This step analyzes each sentence's grammatical structure, linking the dependencies between words and revealing hidden relationships. For example, adhesion analysis can distinguish between biomarkers "associated" with disease and treatments "associated" with disease. This structural analysis is very important for creating meaningful and accurate boundaries in knowledge accounts. Because it ensures that only context-relevant relationships are created, these pre-processing steps transform raw text data into a structured format for knowledge extraction and integration into cognitive neuroscience calculations. This detailed pre-processing pipeline not only improves the quality of the extracted knowledge but also significantly improves the efficiency and accuracy of the subsequent LLM and improves the process to some extent.

### 3.2.Knowledge Graph Representation and Expansion

Cognitive Neuroscience, situated at the intersection of psychology, neuroscience, and computer science, seeks to explore the mysteries of the human mind by examining the neural foundations of cognition[27].

This section describes the formal representation and automatic expansion process of the CNKG using the LLM to generate new entities and relations. This comes from a large collection of cognitive neuroscience literature and is systematically integrated.

CNKG is a structured representation of knowledge in the field of cognitive neuroscience. It is designed to organize and extract meaningful information from large volumes of scientific literature using advanced NLP techniques and LLM such as GPT-4.

CNKG captures relationships between neurons, cognitive functions, and diseases by transforming the original unstructured data into a structured graph format. It consists of nodes representing concepts and specific edges defining connections between them. A key innovation in CNKG design is the ability to extract knowledge and automate graph expansion. So that the framework can predict new relationships between entities(link prediction), which can help identify previously incorrect connections in brain structure and function in advanced research and clinical practice.

CNKG supports tasks such as semantic query and inference. This makes it an invaluable tool for applications such as clinical decision making, researching neurological disorders and the development of new hypotheses about brain function.

Below is a pseudocode representation of the KG expansion process:

```
Algorithm Expand_KG (G, Corpus, LLM, τ)

Input:
  G = (V, E) // Existing KG
  Corpus // Collection of cognitive neuroscience texts
  LLM // Pre-trained Large Language Model (e.g., GPT-4)
  τ // Confidence threshold for adding new relationships

Output:
  G' = (V', E') // Expanded KG

Initialize:
  A = Adjacency_Matrix(G) // Adjacency matrix representing the initial KG

// Step 1: Preprocessing and Tokenization
for each document in Corpus do
  Tokens = Tokenize(document) // Tokenize text into sentences and words
  Entities, Relationships = LLM_Extract(Tokens) // Use LLM to extract entities and relationships

  // Step 2: Entity Recognition and Relationship Extraction
  for each relationship r_ij in Relationships do
    v_i, v_j = Entities[r_ij] // Identify entities involved in the relationship
    P(r_ij) = LLM_Confidence_Score(r_ij) // Get confidence score for the relationship

    // Step 3: Confidence Scoring and Relationship Integration
    if P(r_ij) ≥ τ then
      if v_i not in V then
        Add v_i to V // Add new entity v_i to the vertex set
      end if

      if v_j not in V then
        Add v_j to V // Add new entity v_j to the vertex set
      end if

      // Update adjacency matrix and add new relationship if not already present
      if A[v_i, v_j] == 0 then
        A[v_i, v_j] = 1
        Add e_ij to E // Add new edge (relationship) to the edge set
      end if
    end if
  end for
end for

// Step 4: Entity Addition and Graph Update
return G' = (V', E') // Return the updated KG with expanded entities and relationships

End Algorithm
    return G' = (V', E') with updated A
end algorithm
```

### 3.2.1. Graph Representation (Mathematical Formalism)

We represent the KG, $G = (V, E)$ as a directed graph, where:

$V$ is the set of vertices, representing entities like neurons, cognitive functions, or neurological disorders in cognitive neuroscience. $E$ is the set of directed edges, where an edge $e_{ij} \in E$ defines a directed relationship from vertex $v_i$ to vertex $v_j$, representing causal or associative links between entities. The structure of G is captured using an adjacency matrix $A \in R^{(|V|*|V|)}$, where:

$$A_{ij} = \begin{cases} 1, & \text{if } e_{ij} \in E \\ 0, & \text{otherwise.} \end{cases}$$

$A_{ij}$ is a binary matrix indicating the presence or absence of relationships between entities.

### 3.2.2. Extended Representation with Probability (Confidence-Weighted Matrix)

Incorporating LLM-generated probabilities, we introduce a confidence-weighted matrix $P \in R^{(|V|*|V|)}$, where $P_{ij}$ represents the probability or confidence score of the relationship $e_{ij}$ being valid, as predicted by the LLM:

$$P_{ij} = P(e_{ij} \text{ is valid} | \text{LLM output})$$

This matrix allows for probabilistic reasoning over the graph, enabling decisions about whether to add relationships based on a threshold $\tau$ for the confidence score[28].

### 3.2.3. Automatic Expansion Process (Graph Expansion with Probabilities)

The expansion of the KG using LLMs can be viewed as a probabilistic process where new entities and relationships are integrated into the graph based on their LLM-derived likelihood[29]. The steps for expansion can be formulated mathematically as follows:

1. **Preprocessing and tokenization:** The input corpus $T$ is preprocessed and tokenized into sequences $s_1, s_2, \ldots, s_n$, where each sequence $s_i$ represents a sentence or phrase in the cognitive neuroscience domain. This tokenization ensures that the LLM can parse the text into meaningful entities and relationships.

2. **Entity recognition and relationship extraction:** The LLM is trained to identify entities $v_1, v_2, \ldots, v_k$ and extract relationships $r_{ij}$ between entities $v_i$ and $v_j$ from the textual corpus. Each relationship $r_{ij}$ is accompanied by a confidence score $P(r_{ij})$, representing the probability that the relationship $r_{ij}$ is valid.

3. **Confidence scoring:** For each extracted relationship $r_{ij}$, the LLM assigns a confidence score $P(r_{ij}) \in [0,1]$, which quantifies the model's certainty in the validity of the relationship between entities $v_i$ and $v_j$. The confidence score follows a probabilistic distribution:

$$P_{ij} = P(e_{ij} \text{ is valid} | \text{LLM prediction})$$

This probabilistic score is derived from the LLM's underlying model and reflects how strongly the relationship is supported by the text.

4. **Integration into the graph:** A threshold-based decision rule is used to integrate new relationships into the KG. If the confidence score for a relationship $r_{ij}$ exceeds a predefined threshold $\tau$, the relationship is accepted and added to the graph. The threshold $\tau$ can be interpreted as a risk tolerance parameter, balancing between false positives and false negatives. The adjacency matrix A is updated as:

$$A_{ij} = \begin{cases} 1, & \text{if} P_{ij} \geq \tau \\ 0, & \text{otherwise.} \end{cases}$$

1. This ensures that only relationships with high confidence are added to the graph.

2. **Entity addition:** If the LLM identifies a new entity $v_k$ that does not already exist in $V$, it is added to the vertex set $V$, and any relationships involving $v_k$ are integrated into the adjacency matrix. The graph $G = (V, E)$ is updated as follows:

$$V' = V \cup \{v_k\}$$

$$A' = A \cup \{A_{ik}, A_{kj}\}$$

where new rows and columns are added to the adjacency matrix to accommodate the new relationships involving $v_k$.

### 3.3. Probabilistic Link Prediction and Expansion

Given the structure of the graph $G = (V, E)$ and the confidence-weighted matrix $P$, probabilistic link prediction can be used to suggest new relationships between entities that have not been directly observed but are inferred by the LLM. The probability of a new relationship $r_{ij}$ between entities $v_i$ and $v_j$ can be modeled using a Bayesian framework:

$$P(r_{ij} \text{exists}) = \frac{P(r_{ij})}{1 + P(r_{ij})}$$

This likelihood can be used to predict relationships that are not currently present in the graph but are suggested based on existing patterns and correlations in the data[30].

By combining mathematical probabilistic reasoning with the CNKG expansion process, we build a rigorous framework for expanding KG s with high-confidence entities and relationships. The confidence scores generated by LLM allow probabilistic decisions about which relationships to include in the graph. Ensures the scalability and adaptability of the KG to ensure the graph remains up-to-date with ongoing research in cognitive neuroscience. Below is a pseudocode representation of the knowledge graph expansion process:

Automated expansion of the cognitive neuroscience KG using the ExKG-LLM framework provides a scalable solution for managing and updating knowledge in this rapidly evolving field. This method by systematically integrating new information from scientific texts. It not only improves the integrity and accuracy of CNKG, but also ensures It remains a valuable resource for researchers and clinicians. In this context, the application of LLM exemplifies the power of advanced NLP techniques to transform the way knowledge is structured and accessed in specific scientific domains.

## 4. Evaluation

In this section, we evaluate the effectiveness and efficiency of the extended KG using several key evaluation metrics. These measures help determine the accuracy, completeness, and performance of the knowledge graph after combining new entities and relationships using the LLM. Below is an overview of the criteria and expected results.

### 4.1. Traditional Metrics

The accuracy of the expanded knowledge graph is assessed by comparing the predicted edges $E_{pred}$ with a ground truth set of edges $E_{true}$, using metrics such as precision and recall [31]:

$$Precision = \frac{E_{pred} \cap E_{true}}{E_{pred}}$$

$$Recall = \frac{E_{pred} \cap E_{true}}{E_{true}}$$

### 4.2. Graph Completeness

1. **Vertex Growth:** Assesses the increase in the number of entities (vertices) added to the graph post-expansion. Vertex growth measures the relative increase in the number of vertices (V) after expansion [32]:

$$Vertex_{Growth} = \frac{|V'| - |V|}{|V|} * 100\%$$

Where $|V|$ is the initial number of vertices and $|V'|$ is the number of vertices after expansion.

2. **Edge Growth:** Evaluates the increase in the number of relationships (edges) added to the graph. Edge growth measures the relative increase in the number of edges (E) after expansion:

$$Edge_{Growth} = \frac{|E'| - |E|}{|E|} * 100\%$$

Where $|E|$ is the initial number of edges and $|E'|$ is the number of edges after expansion.

3. **Density:** Measures the ratio of actual edges to possible edges in the graph, indicating how well the graph is connected. Density (D) of a graph is the ratio of the actual number of edges to the maximum possible number of edges between vertices:

$$Density = D = \frac{|E|}{|V|(|V| - 1)}$$

Where $|V|$ is the number of vertices and $|E|$ is the number of edges in the graph.

### 4.3. Graph Consistency

1.  **Conflict detection:** Check for inconsistencies or conflicts in relationships (such as conflicting information about an entities) added by LLM. There is no standard formula for detecting conflicts. But typically newly added edges ($E_{new}$) are compared with existing edges or relationships to identify conflicts. The percentage of conflict (C) can be calculated as[33]:

$$Conflict_{rate} = \frac{|E_{conflict}|}{E_{new}} * 100\%$$

Here $|E_{conflict}|$ is the number of conflicting edges and $E_{new}$ is the total number of newly added edges.

2.  **Validation against ground truth:** New relationships added by the model must be compared with manually collected or validated datasets.

### 4.4. Computational Efficiency

1.  **Time complexity:** it analyzes the time required for knowledge extraction and graph integration processes. It focuses on how LLM scales with larger data sets. The time complexity of an algorithm T(n) is defined as the time required to complete it as a function of the size n of the input. The overall time complexity for the knowledge extraction and integration process can be expressed as follows[34]:

$$T_{total} = O(t_{LLM}) + O(T_{integrate})$$

Where $O(t_{LLM})$ is the time complexity of the LLM and $O(T_{integrate})$ is the time complexity for integrating the extracted knowledge into the graph.

2.  **Space complexity:** It evaluates the memory usage during the expansion process. This is to ensure that this method is feasible for large graphs. The space complexity $S(n)$ specifies the amount of memory space required by the algorithm as a function of the input size $n$ to expand the knowledge calculation:

$$S_{total} = O(S_{LLM}) + O(S_{graph})$$

where $O(S_{LLM})$ is the space complexity of LLM and $O(S_{graph})$ is the space complexity related to graph storage.

**Table 1.** Performance and Structural Comparison: CNKG vs. ExKG-LLM.

| Metric | CNKG | ExKG-LLM | Improvement (%) |
|---|---|---|---|
| Precision | 0.75 | 0.80 | 6.67 |
| Recall | 0.70 | 0.81 | 15.71 |
| F1-Score | 0.72 | 0.805 | 11.81 |
| Node Growth | 3426 | 4150 | 21.13 |
| Edge Growth | 5526 | 7290 | 31.92 |
| Node Types Growth | 11 | 14 | 27.27 |
| Edge Types Growth | 20 | 24 | 20.00 |
| Density | 0.00047 | 0.00042 | -10.64 |
| Conflict Rate (%) | 5% | 6% | 20.00 + |
| Time Complexity | $O(n^2)$ | $O(n \log n)$ | Improved |
| Space Complexity | $O(n)$ | $O(n^2)$ | Less Efficient |

Table 1 shows a significant improvement in the performance of the ExKG-LLM model. Compared with CNKG, the precision, recall and F1 score are improved by 6.67%, 15.71% and 11.81%, respectively, indicating a better precision and balance between precision and recall. The ExKG-LLM approach was able to achieve Precision=0.80, Recall=0.81, F1-score=.80 respectively, while the CNKG approach was able to achieve Precision=0.75, Recall=0.70, F1-score=0.72. A significant increase in node edges was also observed in this approach, with a 21.13% increase in nodes and 31.92% in edges, indicating a more complete and connected graph in ExKG-LLM, but the density decreased by 10.64%, indicating that even If the number of nodes is more and more edges are added, it is more scattered. The rate of conflict increased by 20%, reflecting potential differences in the added relationship. Finally, the time complexity is $O(n^2)$ to $O(n \log n)$, although the space complexity is less efficient. This increases from $O(n)$ to $O(n^2)$, indicating that although the model is faster, it requires more memory.

### 4.4.Link prediction

To evaluate ExKG-LLM, we use three main metrics: MR (average rank), MRR (average mutual rank) and P@K (precision at K), focusing on the link prediction function. These metrics provide us with valuable insights into the graph's performance in predicting relationships between nodes. For filter link prediction tasks We used these three metrics (MR, MRR, and P@K), testing each of them separately[35].

1. **Mean Rank (MR):** Mean Rank is a metric used to evaluate the performance of information retrieval systems. This represents the average order of true positive items in the list of retrieved items. This is calculated by assigning a category to each item. Then find the average of the categories of related items. A lower average rank indicates better performance. This is because related items are rated higher on average.

$$MR = \frac{1}{N} \sum_{i=1}^{N} rank_i$$

Where $rank_i$ is the rank position of the i-th relevant item, and $N$ is the total number of relevant items.

2. **Mean Reciprocal Rank (MRR)**: Mean Reciprocal Rank is a metric that can be understood with a helpful example. Imagine you are searching for a specific document in a large database. MRR is the average of the reciprocal rows of the first related item in the list of retrieved items. This is useful in situations where the first relevant outcome has priority. MRR is a measure of how quickly the first relevant object is retrieved[36].

$$MRR = \frac{1}{N} \sum_{i=1}^{N} \frac{1}{rank_i}$$

Where $rank_i$ is the rank position of the first relevant item for the i-th query, and $N$ is the total number of queries.

3. **Precision in K (P@K)**: P@K measures the proportion of the top K items retrieved to their corresponding items. In our evaluation, we calculate P@K for different values of K, specifically K=1, K=3, and K=10. P@K measures the proportion of relevant items in the first k results, it evaluates the quality of the results[36]. The highest ranking and calculated as follows:

$$P@K = \frac{1}{N} \sum_{i=1}^{N} \frac{\text{Number of relevant items in top K}}{K}$$

where $K$ is the number of top items considered, and N is the total number of queries.

We employ link prediction methods such as TransE, RotatE, DistMult, ComplEx, ConvE , HolmE which leverage advanced algorithms to infer connections within the graph.

**Table 2.** Comparison of Link Prediction for ExKG-LLM with TransE, RotatE, DistMult, ComplEx, ConvE, and HolmE using MR, MRR, and P@K Metrics with Other KGs

| KG | Metric | TransE [40] | RotatE [41] | DistMult [42] | ComplEx [43] | ConvE [44] | HolmE [45] |
|---|---|---|---|---|---|---|---|
| FB15k | MR | 45 | 42 | 173 | 33 | 51 | - |
| | MRR | 0.628 | 0.791 | 0.784 | 0.855 | 0.688 | - |
| | P@1 | 0.493 | 0.739 | 0.736 | 0.823 | 0.594 | - |
| | P@3 | 0.578 | 0.788 | 0.733 | 0.759 | 0.801 | - |
| | P@10 | 0.847 | 0.881 | 0.863 | 0.91 | 0.849 | - |
| FB15k-237 | MR | 209 | 178 | 199 | 144 | 281 | - |
| | MRR | 0.31 | 0.336 | 0.313 | 0.367 | 0.305 | 0.331 |
| | P@1 | 0.217 | 0.238 | 0.224 | 0.271 | 0.219 | 0.237 |
| | P@3 | 0.257 | 0.328 | 0.263 | 0.275 | 0.35 | 0.366 |
| | P@10 | 0.496 | 0.53 | 0.49 | 0.558 | 0.476 | 0.517 |
| WN18 | MR | 279 | 274 | 675 | 190 | 413 | - |
| | MRR | 0.646 | 0.949 | 0.824 | 0.951 | 0.945 | - |
| | P@1 | 0.405 | 0.943 | 0.726 | 0.944 | 0.938 | - |
| | P@3 | 0.888 | 0.953 | 0.914 | 0.945 | 0.947 | - |
| | P@10 | 0.948 | 0.96 | 0.946 | 0.961 | 0.956 | - |
| WN18RR | MR | 3936 | 3318 | 5913 | 2867 | 4944 | - |
| | MRR | 0.206 | 0.475 | 0.433 | 0.489 | 0.427 | 0.466 |
| | P@1 | 0.279 | 0.426 | 0.396 | 0.442 | 0.389 | 0.415 |
| | P@3 | 0.364 | 0.492 | 0.44 | 0.46 | 0.43 | 0.489 |
| | P@10 | 0.495 | 0.573 | 0.502 | 0.58 | 0.507 | 0.561 |
| YAGO3-10 | MR | 1187 | 1830 | 1107 | 793 | 2429 | - |
| | MRR | 0.501 | 0.498 | 0.501 | 0.577 | 0.488 | 0.441 |
| | P@1 | 0.405 | 0.405 | 0.412 | 0.5 | 0.399 | 0.333 |
| | P@3 | 0.528 | 0.55 | 0.38 | 0.4 | 0.56 | 0.507 |
| | P@10 | 0.673 | 0.67 | 0.661 | 0.7129 | 0.657 | 0.641 |
| CNKG | MR | 87 | 97 | 308 | 109 | 102 | 277 |
| | MRR | 0.302 | 0.251 | 0.51 | 0.221 | 0.231 | 0.19 |
| | P@1 | 0.31 | 0.256 | 0.251 | 0.274 | 0.21 | 0.065 |
| | P@3 | 0.494 | 0.309 | 0.401 | 0.311 | 0.294 | 0.089 |
| | P@10 | 0.571 | 0.35 | 0.652 | 0.325 | 0.319 | 0.121 |
| ExKG-LLM | MR | 144 | 175 | 562 | 101 | 259 | 304 |
| | MRR | 0.526 | 0.311 | 0.198 | 0.198 | 0.396 | 0.181 |
| | P@1 | 0.575 | 0.366 | 0.274 | 0.231 | 0.412 | 0.049 |
| | P@3 | 0.602 | 0.415 | 0.489 | 0.282 | 0.463 | 0.075 |
| | P@10 | 0.644 | 0.462 | 0.727 | 0.327 | 0.513 | 0.104 |

Table 2 compares several knowledge base embedding models, including TransE, RotatE, DistMult, ComplEx, ConvE, and HolmE, of which ExKG-LLM is the proposed model of this study. Considering the MR (mean rank) values, ExKG-LLM shows competitive performance, especially FB15k -237 and WN18RR on datasets outperforming models such as DistMult and ConvE. However, it has problems with metrics like P@1 and MRR. For some datasets, such as WN18 and YAGO3-10, traditional models such as ComplEx and RotatE perform significantly better. In particular, ExKG -LLM shows strong performance at P@3 and P@10 for datasets such as FB15k and FB15k-237, demonstrating its ability to accurately predict the top 3 to 10 results. In general, ExKG-LLM performs better in some cases and it is more acceptable that this is due to some key indicators behind the curtain of this model. This model was able to achieve a value of 101 in the best MR criterion in ComplEx, the best MRR criterion in HolmE. The highest values of P@K were respectively for P@1=.575, P@3=0.602, and P@10=0.644 in TransE.

In the MR chart (fig1), CNKG consistently has lower (better) mean ranks compared to ExKG-LLM across most models. The only exception is the ComplEx model, where ExKG-LLM shows a lower MR. This suggests that CNKG generally performs better at placing the correct results closer to the top of the ranking.

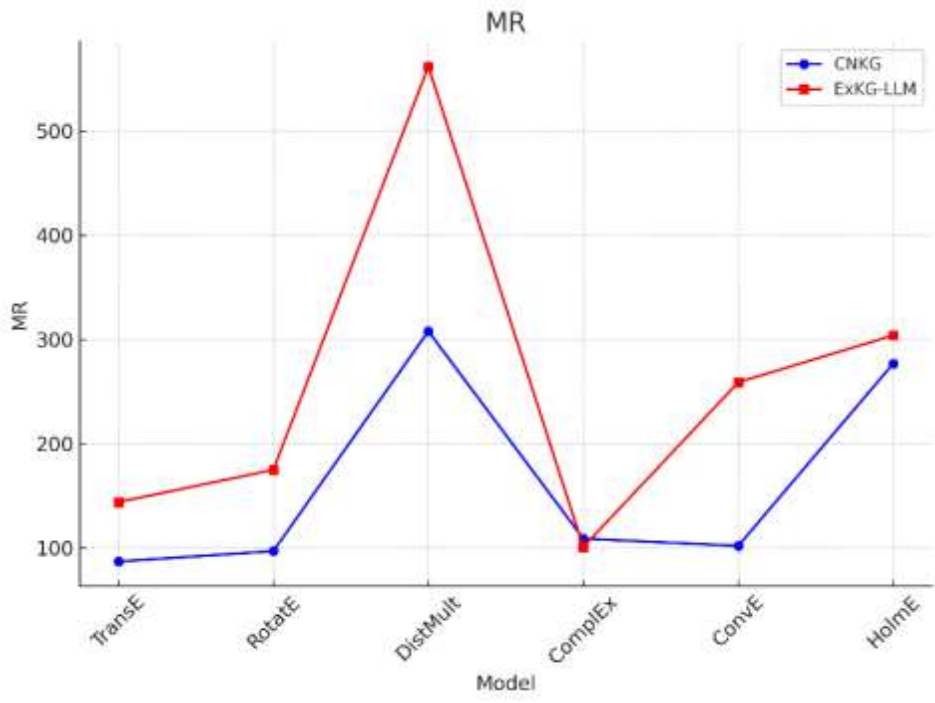

**Figure 1.** Comparison of MR between CNKG and ExKG-LLM across models

The MRR chart (fig2) shows that ExKG-LLM has higher MRR values for TransE and RotatE models, indicating better overall ranking precision in these models. However, CNKG performs significantly better in the DistMult model, showcasing its strength in that model compared to ExKG-LLM.

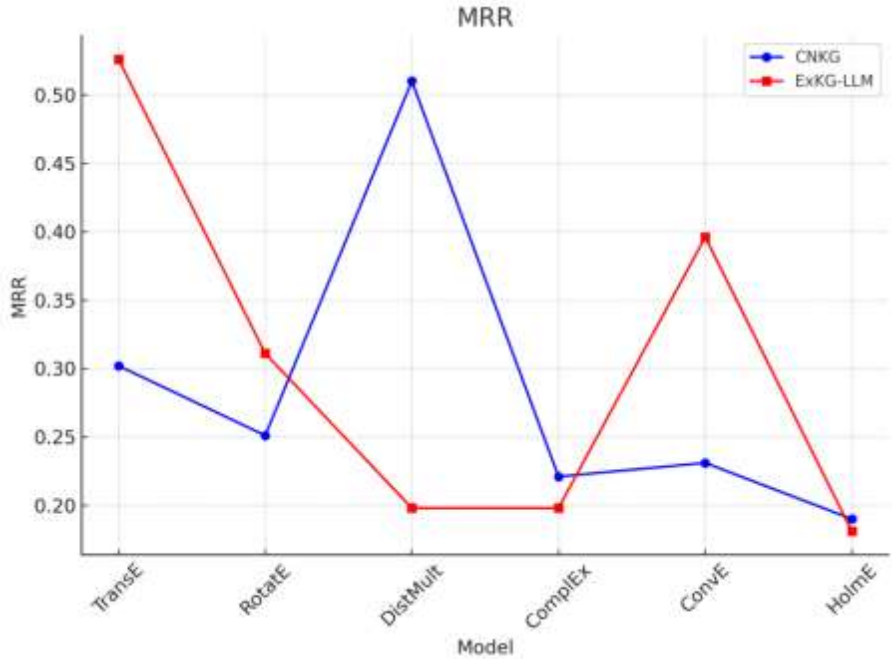

**Figure 2.** Comparison of MRR between CNKG and ExKG-LLM across models

For P@1 (fig3), ExKG-LLM outperforms CNKG in most models, especially in TransE and ConvE, showing that it is better at retrieving the correct result as the top prediction. CNKG, however, performs better in the DistMult model.

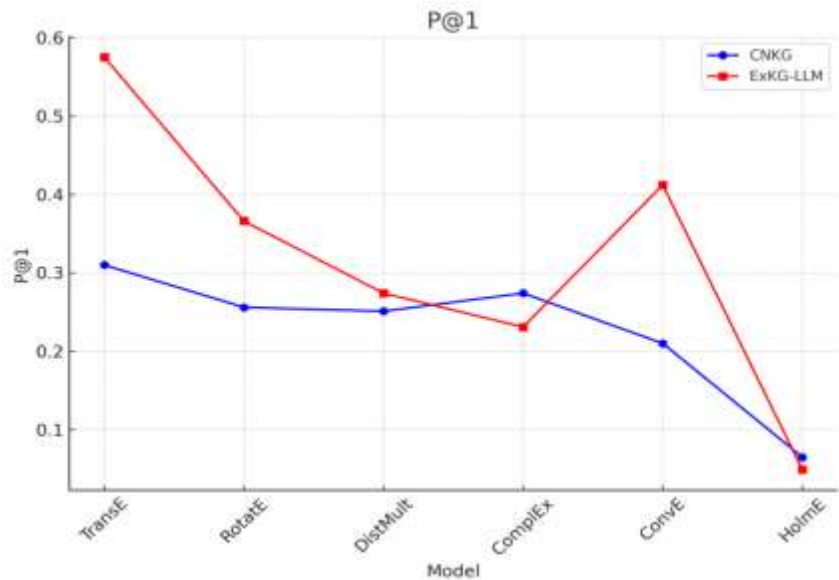

**Figure 3.** Comparison of P@1 between CNKG and ExKG-LLM across models

The P@3 (fig4) chart demonstrates a similar pattern to P@1, where ExKG-LLM generally performs better, particularly in the RotatE and TransE models. CNKG again shows stronger performance in DistMult, indicating that CNKG's top 3 predictions are more accurate in this model.

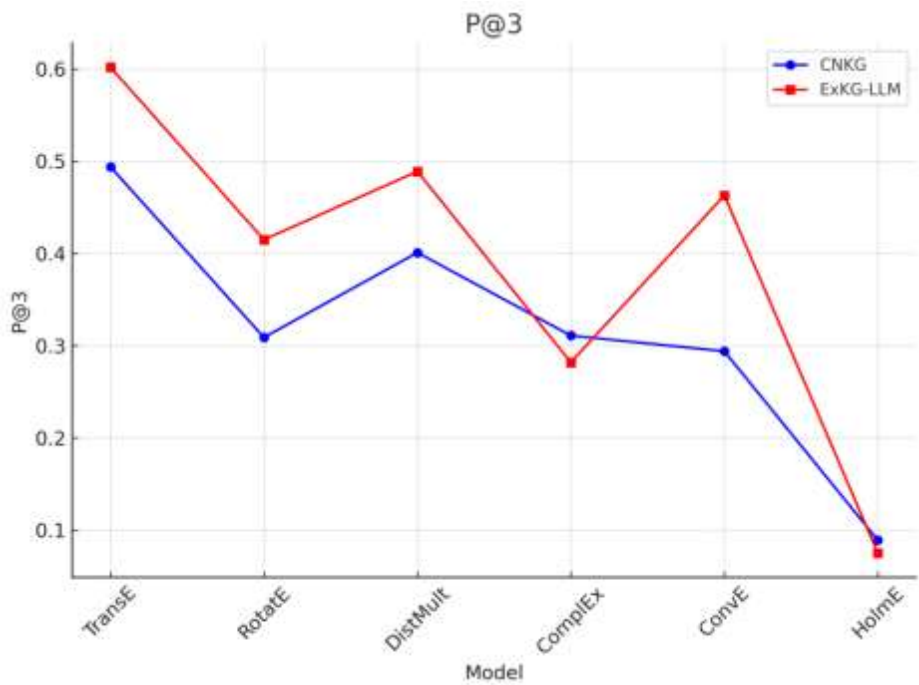

**Figure 4.** Comparison of P@3 between CNKG and ExKG-LLM across models

In the P@10 (fig5) chart, both CNKG and ExKG-LLM show competitive performance, with ExKG-LLM slightly outperforming CNKG in RotatE and ConvE models. However, CNKG has a distinct advantage in DistMult, suggesting it provides better quality results within the top 10 for this model.

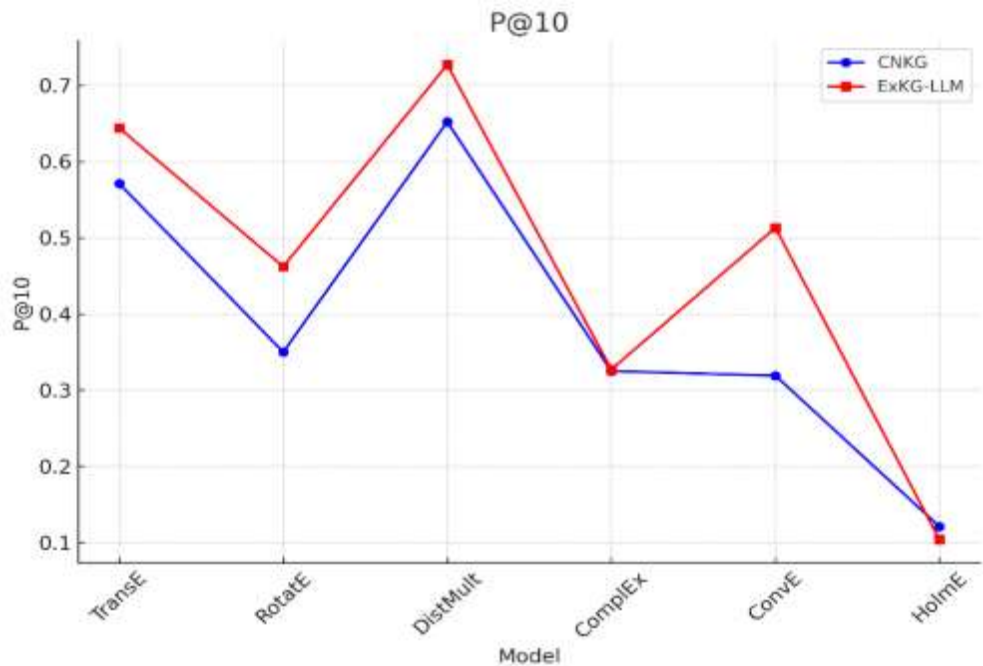

**Figure 5.** Comparison of P@10 between CNKG and ExKG-LLM across models

### 4.5. Expert Review

**Named Entity Recognition (NER)** The task of Named Entity Recognition (NER) involves identifying and classifying key entities such as diseases, medications, symptoms, and body parts from unstructured medical texts. Typically, NER in medical texts works as follows[46] :

1. **Input:** A medical text containing descriptive data, test results, clinical records, etc.

2. **Output:** Identification of specific entities like disease names (e.g., "heart failure"), medication names (e.g., "Captopril"), or symptoms (e.g., "chest pain").

**Relation Extraction (RE)** Relation Extraction (RE) focuses on identifying the relationships between named entities within a text. For example, a relationship can be that between drug interactions, a symptom with a disease, or an effect of a drug regarding a particular condition[47].

1. **Input:** A medical text containing two entities such as "Captopril" and "heart failure."

2. **Output:** The relationship between these entities (e.g., "Captopril is an ACE inhibitor used to treat heart failure").

**Evaluation and Calculation by Experts** To assess the efficacy models in NER and RE tasks, annotated medical data labelled by experts are usually utilized. Such a process is carried out as follows:

1. Creation of Annotated Datasets: Medical experts review the textual data and manually label key entities and relationships between them. These annotations serve as the ground truth for model evaluation.

2. Evaluation Metrics:

    1. **Precision:** Precision is the ratio of correctly identified entities or relations by the model compared with the total number of entities or relations identified by it. Thus it defines the accuracy of the predictions made by the model.

    $$Pricsion = \frac{True_{Positives}}{True_{Positives} + False_{Positives}}$$

    2. **Recall:** Recognized to be the correct ratio between the actual identified entities or relations and the total number of entities or relations present in the data. In essence, the metric captures the relative completeness of the model about relevant information.

    $$Pricsion = \frac{True_{Positives}}{True_{Positives} + False_{Negatives}}$$

    3. **F1 measure:** F1 measure is the calculated harmonic mean of precision and recall. It is a "fair" assessment of both measures, proving useful when precision and recall need to be considered together.

The results generated after training the model will be compared with the manual annotations, and hence the performance would be based on precision, recall, and F1-Score. Also, they check the suitability of the model to work with Out-of-Distribution (OOD) data and decide whether the model correctly recognizes entities or relationships unseen during training. Named Entity Recognition (NER) as well as Relation Extraction (RE) are some major tasks carried out on medical text, allowing LLMs to extract useful data from the unstructured data source. The best way to evaluate such models is through datasets manually labeled along with evaluation parameters like precision, recall, and F1-Score. This is so very important in some medical applications, such as disease diagnosis and drug prescription, where good and accurate information extraction is very necessary.

**Table 3.** RE (ExKG-LLM, StrokeKG)

| Metric | Precision | Recall | F1 |
|---|---|---|---|
| ExKG-LLM (Expert Review) | 81.73 | 90.27 | 85.79 |
| StrokeKG [48] | 80.06 | 88.92 | 84.26 |

**Table 4.** NER(ExKG-LLM, StrokeKG, Heart Failure KG )

| Metric | Precision | Recall | F1 |
|---|---|---|---|
| ExKG-LLM (Expert Review) | 89.46 | 92.21 | 90.81 |
| StrokeKG | 94.21 | 86.04 | 90.26 |
| Heart Failure KG [49] (TwoStepChat-zeroshot) | 82.33 | 88.50 | 85.31 |
| Heart Failure KG (TwoStepChat-fewshot 10) | 87.35 | 91.35 | 89.31 |

### 4.6. Complex Network

KG was considered a complex network and was further analyzed to understand the characteristics and behavior of the structure. This approach allows for a deeper exploration of the connections and patterns within ExKG-LLM and provides insight into its complex nature and potential impacts.

Using Gefi software[50], we can calculate two main parameters of KG: average clustering coefficient and diameter. A measure of node closeness between nodes in a network. The average cluster coefficient can indicate the existence of interconnected communities or clusters. Longest diameter or shortest path It represents the maximum shortest path length between any pair of nodes in the network and provides an overview of the overall connectivity of the network or the distance between the farthest nodes.

**Table 5.** Comparison of ExKG-LLM with other complex networks in terms of Average Clustering Coefficient and Diameter

| Name | # Nodes | # Edges | Average clustering coefficient | Diameter |
|---|---|---|---|---|
| CC-Neuron [51] | 1018524 | 24735503 | 0.143218 | 15 |
| ChCh-Miner [51] | 1514 | 48514 | 0.303968 | 7 |
| PP-Decagon [51] | 19081 | 715612 | 0.233544 | 8 |
| DD-Miner [51] | 6878 | 6877 | 0.000000 | 23 |
| GrGr-EnhancedHiC5K[51] | 1419 | 43763 | 0.867585 | 10 |
| CNKG | 3426 | 5526 | 0.419541 | 13 |
| ExKG-LLM | 4150 | 7290 | 0.420027 | 15 |

The ExKG-LLM shows improvements over CNKG in both average clustering coefficient and diameter. With a clustering coefficient of 0.420027, ExKG-LLM slightly surpasses CNKG's 0.419541, reflecting marginally stronger local clustering. However, in terms of diameter, ExKG-LLM has a larger diameter of 15 compared to CNKG's 13, suggesting that while ExKG-LLM is more expansive with more nodes and edges, CNKG is slightly more compact and efficient in terms of node connectivity.

When compared to other complex networks, ExKG-LLM exhibits stronger clustering than CC-Neuron (0.143218) and PP-Decagon (0.233544) but is still outperformed by GrGr-EnhancedHiC5K (0.867585), which shows the highest clustering among the networks. In terms of diameter, ExKG-LLM's diameter of 15 is comparable to CC-Neuron but larger than most other networks except DD-Miner, which has the largest diameter of 23. This makes ExKG-LLM more interconnected than many other networks, though it remains more expansive in its structure than CNKG.

In the first chart, the Average Clustering Coefficient of different complex networks is compared. CC-Neuron has the lowest value at 0.143218, indicating weak clustering tendencies. ChCh-Miner and PP-Decagon have higher values of 0.303968 and 0.233544, respectively, showing moderate clustering. GrGr-EnhancedHiC5K stands out with a very high clustering coefficient of 0.867585, indicating extremely tight clusters. DD-Miner has a clustering coefficient of 0, suggesting no clustering at all. Focusing on CNKG (0.419541) and ExKG-LLM (0.420027), their values are quite close, indicating similar levels of clustering in both networks, though they have higher clustering than most others except GrGr-EnhancedHiC5K.

In the second chart, the Diameter of the networks is analyzed. DD-Miner has the largest diameter at 23, meaning the nodes are very spread out. CC-Neuron has a diameter of 15, the same as ExKG-LLM, showing comparable network spread. ChCh-Miner has the smallest diameter at 7, indicating a more compact structure. PP-Decagon and GrGr-EnhancedHiC5K have diameters of 8 and 10, respectively, showing moderate spread. CNKG, with a diameter of 13, is somewhat more compact than ExKG-LLM (15), though both have similar diameters to CC-Neuron, suggesting that these networks have a balance between node connectivity and network size.

## 5. Conclusion and Future work

The ExKG-LLM framework has shown significant potential in automatically expanding the Neuroscience CNKG using LLMs. The framework's ability to fully improve graphs, precision, overall connectivity through the integration of new entities and relationships is a valuable resource for improving key metrics such as precision, recall, and F1 score, along with the dramatic growth of edge nodes, highlighting the benefits of ExKG-LLM for It is the progress and expansion of knowledge.

Despite these advances, challenges remain, with a 20\% increase in conflict rates with conflicting relationships indicating the need for more sophisticated verification mechanisms to ensure consistency. Also, reducing the density of the graph shows that as more nodes and edges are added, their distribution is scattered and may limit the usefulness of the graph, which will be necessary for customization.

Considering complex network analysis, ExKG-LLM shows strong local clustering. CNKG has increased in diameter (with a clustering coefficient of 0.420027) but has also increased in diameter by up to 15, indicating a more detailed but possibly less connected structure compared to CNKG.

Link prediction is another important area for future development, although ExKG-LLM integrates new knowledge. effectively Incorporating improved link prediction algorithms can help improve the ability to infer unknown connections between entities. Techniques such as TransE, RotatE, ConvE can be used to improve the prediction accuracy of graphs. Scalability is another important area for improvement. While the time complexity of ExKG-LLM was optimized from O($n^2$) to O($n \log n$), the space complexity increased from O(n) to O($n^2$), maintaining or Improving processing speed will be important to freeing up memory. The performance of ExKG-LLM on large datasets in more complex areas is used.

Additionally, the adaptability of the framework to other scientific fields should be examined. Extensions of the ExKG-LLM methodology will examine generalizability across domains such as genomics, pharmacology, or other interdisciplinary areas. Considering the complex relationships in these areas Investigating how ExKG-LLM performs in different scientific contexts It will help determine the ability to be used more broadly.